\def\year{2019}\relax
\documentclass[letterpaper]{article} 
\usepackage{aaai19}  
\usepackage{times}  
\usepackage{helvet}  
\usepackage{courier}  
\usepackage{url}  
\usepackage{graphicx}  

\usepackage{subfigure}
\usepackage{multirow}

\title{Improving Safety in Reinforcement Learning Using Model-Based Architectures and Human Intervention\thanks{This work is supported
by U.S. Army Research Laboratory under Cooperative Agreement Number W911NF-10-2-0022}}
\author{
Bharat Prakash\textsuperscript{\rm 1}\thanks{Denotes equal contribution},
Mohit Khatwani\textsuperscript{\rm 1}\footnotemark[2],
Nicholas Waytowich\textsuperscript{\rm 2},
Tinoosh Mohsenin\textsuperscript{\rm 1},\\
\textsuperscript{\rm 1}University of Maryland, Baltimore County,
\textsuperscript{\rm 2} US Army Research Laboratory\\
bhp1@umbc.edu, khatwan1@umbc.edu, nicholas.r.waytowich.civ@mail.mil, tinoosh@umbc.edu
}

\begin{document}
\maketitle
\begin{abstract}
Recent progress in AI and Reinforcement learning has shown great success in solving complex problems with high dimensional state spaces. However, most of these successes have been primarily in simulated environments where failure is of little or no consequence. Most real-world applications, however, require training solutions that are safe to operate as catastrophic failures are inadmissible especially when there is human interaction involved. Currently, Safe RL systems use human oversight during training and exploration in order to make sure the RL agent does not go into a catastrophic state. These methods require a large amount of human labor and it is very difficult to scale up. We present a hybrid method for reducing the human intervention time by combining model-based approaches and training a supervised learner to improve sample efficiency while also ensuring safety. We evaluate these methods on various grid-world environments using both standard and visual representations and show that our approach achieves better performance in terms of sample efficiency, number of catastrophic states reached as well as overall task performance compared to traditional model-free approaches
\end{abstract}

\section{Introduction}

Recent progress in AI and Reinforcement Learning (RL) has shown success in learning policies to solve complex tasks such as playing video games from images \cite{mnih2015human}, or robotic maneuvering and manipulation \cite{schulman2015trust}. However, most of these successes were achieved in simulated environments where unsupervised exploration during training is amenable as failure states are of little consequence to the learning agent and it's surroundings. Most real-world applications, which require training to be done in-situ, will require the agent or robot to act safely while learning. In this case, completely unsupervised exploration during training, which lead to catastrophic failures are inadmissible and an approach for safe learning is required, especially during the initial exploration phases.

This has encouraged a new sub branch of reinforcement learning called Safe RL. Although there are different ways of achieving safety during RL, most of the successful Safe RL systems use human oversight during training and exploration in order to make sure the AI agent does not go into a catastrophic state \cite{Saunders2017}, and avoid potential damage to property or any humans involved. Some notable examples are self driving cars and drones which almost always use a human who oversees the actions of the agent and can intervene if necessary.

A downside to these human-in-the-loop, safe RL techniques is that they often require a lot of human time and do not scale up very well due to the relatively poor sample-efficiency inherit to RL. For many complicated tasks with high-dimensional state-spaces, it can easily become infeasible (in terms of human labor required) to train these models safely. Some previous research \cite{Saunders2017} show ways of reduce human time by training a supervised  learner to imitate human intervention and avoid catastrophes. Such methods do help, but they are not very data efficient and can still require a lot of human time before the supervised learners can take over.

In this paper, we present a method for improving these schemes using model-based RL and show how they can improve sample efficiency compared to existing safe RL methods. We present a hybrid scheme to train RL agents in a safe manner with minimum human intervention time. We use a model based approach where we learn the dynamics of the environment and a Model Predictive Controller (MPC) to initialize the policy of the model free agent \cite{richards2005robust}. We also train a blocker agent which is a supervised learner that is trained to imitate a human overseer and block unsafe actions. We show that this hybrid approach requires less human intervention time while achieving the same or better performance in terms of rewards and safety compared pure model free systems. Using two safe-RL environments (GridWorld and Island Navigation) we show that compared to traditional policy gradient approaches, our hybrid model achieves 5$\times$ reduction in number of catastrophic states encountered. Furthermore, we show that our approach is more sample efficient than traditional model-free approaches for safe-RL, obtaining higher task performance in significantly less training time.

\section{Related Work}
There are various ways in which human input can be used to augment or improve the training of a learning agent. The most common approach involves using human provided demonstrations of a given task and using imitation learning to directly clone or imitate the demonstrated behaviour \cite{hussein2017imitation}. However, imitation learning cannot be applied in cases where it is difficult for the human to perform the task well (or even at all). \citeauthor{christiano2017deep} shows another method where human feedback (in the form of preferences) can be used to learn a reward function for an RL agent \cite{christiano2017deep}. In work from \citeauthor{Warnell2018}, the reward function is learned directly from scalar valued human feedback during the learning process \cite{Warnell2018}. Recently, an approach from \citeauthor{Waytowich2018} combined multiple forms of human interaction to train AI agents safely by combining learning from human demonstrations and learning from human interventions \cite{Waytowich2018,Goecks2018}. All of these methods, however, are model-free and thus can still suffer from poor sample efficiency.

Recently, \citeauthor{nagabandi2018neural} show how model-based algorithms can be used for efficient learning due to their low sample requirements \cite{nagabandi2018neural}. In this work, they initialize a policy gradient method using good trajectories observed in the model based training of the agent which improves the sample efficiency of the learned policy.

Additionally, \citeauthor{Saunders2017} show a way to formalize this human intervention and help the AI agents to learn safely during training \cite{Saunders2017}. In this approach, they train an agent to act as a blocker in a supervised approach to imitate the human (who initially acts as the blocker) and intervene when the RL agent is about to take an action which can lead to a catastrophic state. Data is collected to train this blocker during the human oversight phase. Despite this, \citeauthor{Saunders2017} assert that since the training of this blocker needs large amounts of high quality data, the amount of human oversight time required can get unfeasible with more complex problems. Our approach seeks to directly overcome this challenge by introducing model-based learning into this model-free safe-RL approach to improve sample efficiently and reduce the amount of human oversight required.

\begin{figure}
		\centering{\includegraphics[width=.95\columnwidth]{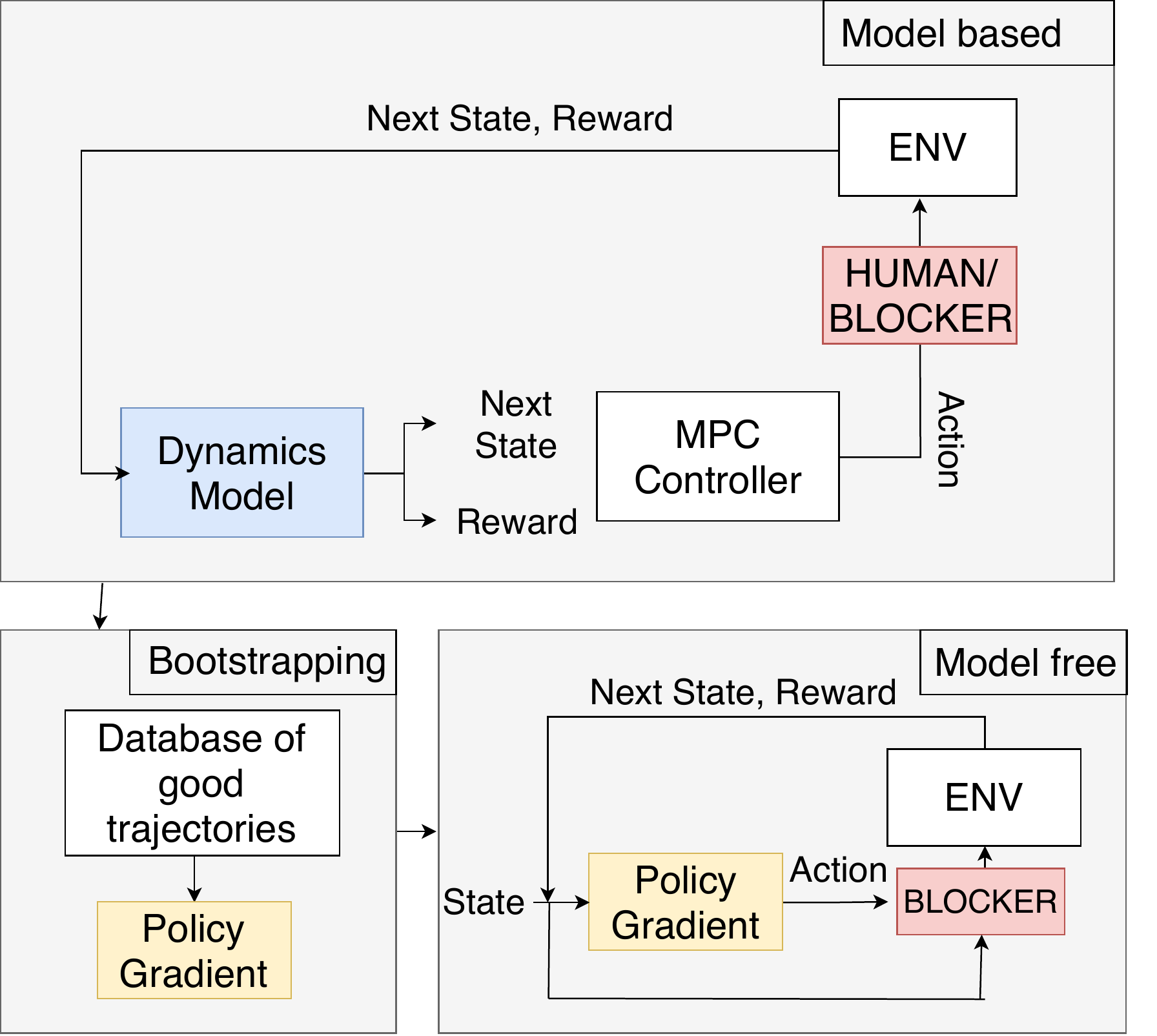}}
		\caption{Architecture which mainly consists of three blocks model based, bootstrap and model free. Blue colored block is the dynamics model which tries to imitate the real environment. Red block denoted the human or blocker (human imitator)}
		\label{arch}
\end{figure}

\section{Methods}
We now present our architecture for safe reinforcement learning using hybrid model-based and model-free techniques.
Our architecture, which is outlined in Figure~\ref{arch}, includes three main modules, a model based module, a bootstrapping module and a model free module. First, the model-based system consists of a dynamics model that drives an MPC controller which is supervised by either the human or a learned blocker agent to prevent catastrophic actions.
Second, the bootstrapping module takes high-quality examples generated from the MPC to initialize a model-free RL algorithm. Finally the model-free module uses a bootstrapped policy-gradient based RL agent to continue learning the task under the supervision of the blocker agent.

\subsection{Blocker Agent}
There are various ways in which safety can be ensured during training. One such way involves using a human in the training phase to intervene (or block actions) so that agent doesn't take actions which lead to catastrophic states. \cite{Saunders2017} introduce a way to train a supervised learner which can be trained to imitate a human and block actions that are unsafe.

We follow a similar method where we built a web interface which allows users to monitor agents and block unsafe actions. If an action is blocked, the agent is forced to select another action. The data is collected and and eventually a model is trained to perform this task of blocking unsafe actions.

This reduces the human labor time considerably and makes this process somewhat feasible. However, the amount of data required to train a good blocker agent can still be large. Our method seeks to improve the blocker performance and reduce the amount of training data required by using a model-based policy for generating the training dataset for the blocker. This dataset is collected during the initial exploration phase of the whole system (described in more detail below).

\subsection{Hybrid Model-based Reinforcement Learning}
Typically, model-based systems attempt to learn a dynamics model of the environment so that this learned model can then be used in various ways to improve the learning of a policy.
In our approach, we learn a dynamics model that will then be used to select actions to be taken in an environment using a model-predictive controller (MPC). We start by training our dynamics model with random exploration of the environment for 50 episodes.
After this pre-training stage, our dynamics model is used to drive an MPC controller and ran for 150 episodes, during which, the data is used to further improve the dynamics model.  The MPC controller we used in our experiments is a simple random shooting method \cite{richards2005robust} where $K$ random action trajectories are generated each with horizon $H$. These random action trajectories are then evaluated, the trajectory having maximum overall reward is chosen and the first action from that trajectory is executed.

During the 150 episodes with the MPC controller, we select successful trajectories (where the agent successfully reached the goal state) and store them into database buffer in the bootstrapping module, which are used to boot-strap a policy gradient model.  After 150 episodes of the MPC controller are completed, we switch to the model-free module which takes the boot-strapped RL agent and continues to learn the task with the trained blocker agent for 1000 episodes. We use the REINFORCE policy gradient algorithm for our model-free RL agent \cite{sutton2000policy}.

During this entire training cycle, we also have the human/blocker agent which intervenes and blocks unsafe actions. The human is used to block actions for the first 25 episodes (up to 1000 steps) during which the data generated is used to train the blocker. After 1000 steps, the human is replaced by the blocker agent for the remainder of the training cycle.

\subsection{Model Architectures}
\begin{figure}[t]
		\centering{\includegraphics[width=0.95\columnwidth]{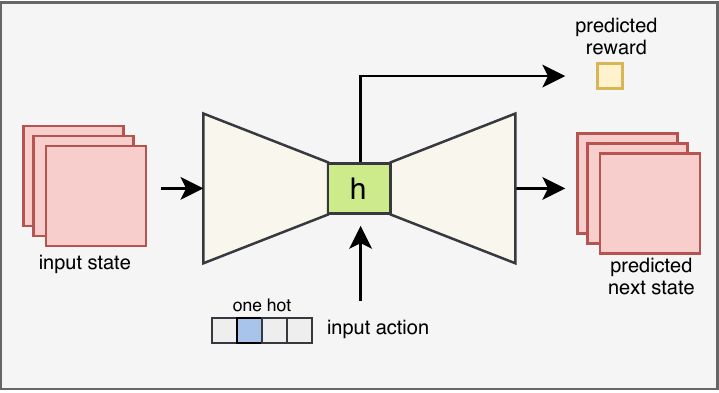}}
		\caption{Auto-encoder architecture for the Island Navigation task. The input state is passed through several convolutional layers. The input action is appended to the encoded state representation, and then the decoder is used to output the predicted reward as well as the predicted next state}
		\label{ae_arch}
\end{figure}

The dynamics model, which is shown in the blue box in Figure \ref{ae_arch}, is a deep neural network which takes the current state and action as input and predicts the next state and immediate reward.

\begin{figure}
  \centering
  \subfigure[4x4 GridWorld]{\includegraphics[scale=0.3]{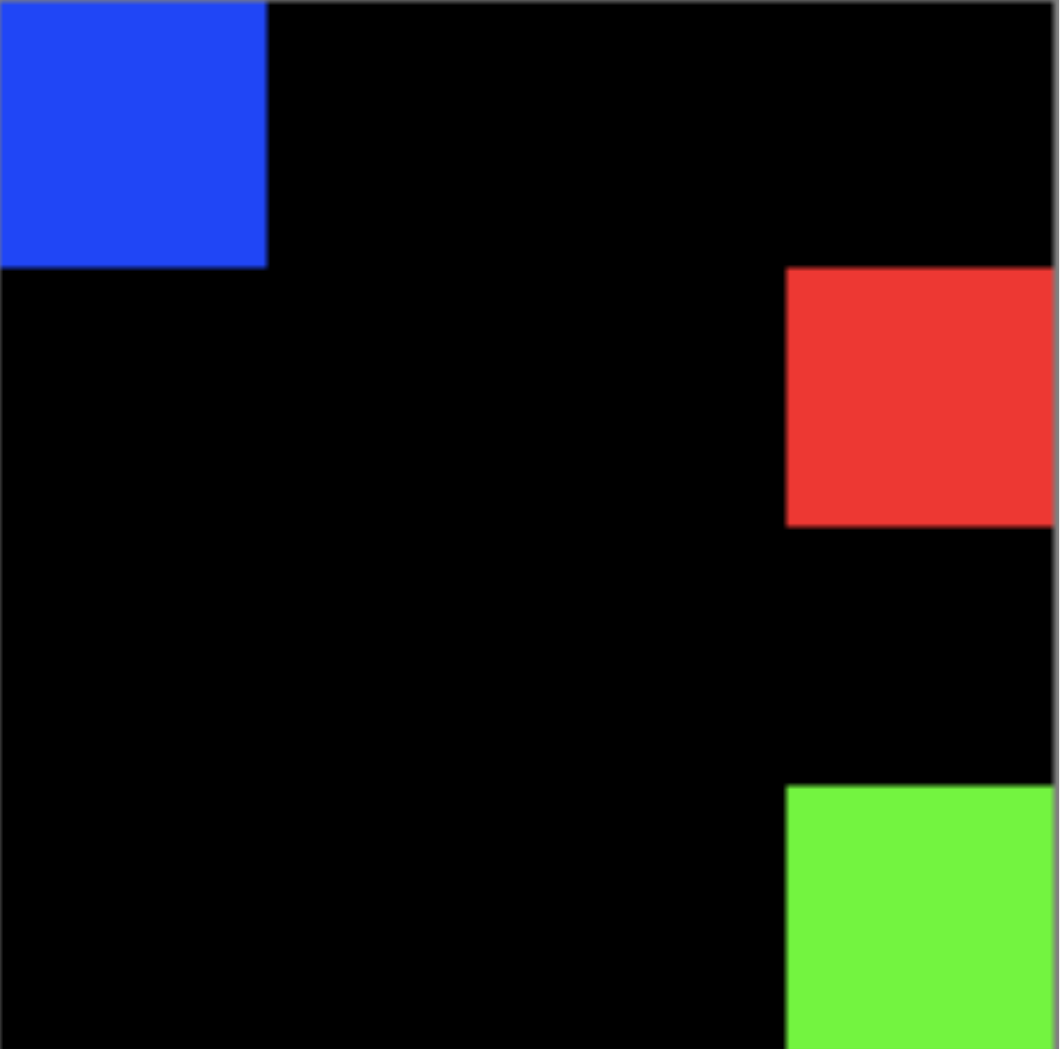}}\quad
  \subfigure[Island Navigation]{\includegraphics[scale=0.4]{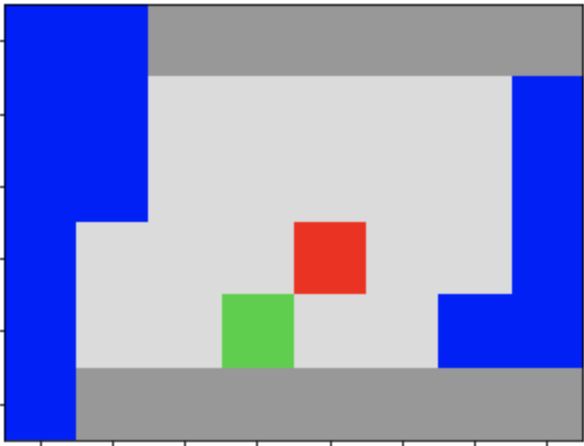}}\quad
    \caption{Environments for experiments. (a) is in standard representation from OpenAI Gym implementation of text based grid-world. (b) has states in visual representation from Deepmind's AI safety grid-worlds}
    \label{fig:env}
\end{figure}

In the 4x4 grid-world environment, shown in Figure \ref{fig:env} (a), we use a standard representation and a simple feed forward neural network. Here we concatenate the state and action which acts as the input and we predict the next state and reward. Our dynamics model for 4x4 grid-world with standard representation consists of two fully connected layers with 32 and 16 neurons respectively. It uses a ReLU activation function after each dense layer. We optimize using a categorical cross-entropy loss function and Adam optimizer with learning rate of 0.001

For Island Navigation, shown in Figure \ref{fig:env}(b), we use visual representations (i.e. learn from images). Since the input is a 32x32 sized image, we use a CNN auto-encoder to predict the next state. In this case, we append the action to the encoded state representation and the output from the decoder side is the next state which is again a 32x32 image and a scalar reward (shown in Figure \ref{ae_arch}). Details of the convolutional neural network architecture are listed in Table \ref{fig:ae_cnn_arch}. This auto-encoder was trained using categorical cross entropy loss and RMSProp opmtimizer with learning rate as 0.001.

\begin{table}[t]
\small
\begin{tabular}{|l|l|l|l|l|}
\hline
\textbf{Layer} & \textbf{\begin{tabular}[c]{@{}l@{}}input\\ channel/\\ features\end{tabular}} & \textbf{\begin{tabular}[c]{@{}l@{}}output\\ channel/\\ features\end{tabular}} & \textbf{\begin{tabular}[c]{@{}l@{}}kernel \\ size\end{tabular}} & \textbf{activation} \\ \hline
\textbf{conv1} & 3 & 3 & (3,3) & ReLU \\ \hline
\textbf{conv2} & 3 & 32 & (2,2) & ReLU \\ \hline
\textbf{conv3} & 32 & 32 & (3,3) & ReLU \\ \hline
\textbf{conv4} & 32 & 32 & (3,3) & ReLU \\ \hline
\textbf{fc1} & 8192 & 128 &  & ReLU \\ \hline
\textbf{fc2} & 128 & 24 &  &  \\ \hline
\textbf{fc3} & \begin{tabular}[c]{@{}l@{}}24 + \\ 4 (action)\end{tabular} & 128 &  & ReLU \\ \hline
\textbf{fc4} & 128 & 8192 &  & ReLU \\ \hline
\textbf{\begin{tabular}[c]{@{}l@{}}fc4\\ (reward\\ output)\end{tabular}} & 128 & 3 &  & Softmax \\ \hline
\textbf{deconv1} & 32 & 32 & (3,3) & ReLU \\ \hline
\textbf{deconv2} & 32 & 32 & (3,3) & ReLU \\ \hline
\textbf{deconv3} & 32 & 32 & (2,2) & ReLU \\ \hline
\textbf{\begin{tabular}[c]{@{}l@{}}conv5\\ (output)\end{tabular}} & 32 & 3 & (3,3) & Sigmoid \\ \hline
\end{tabular}
\caption{Auto-encoder CNN architecture details. fc is for fully connected layers.}
\label{fig:ae_cnn_arch}
\end{table}

\begin{table}
\resizebox{.95\columnwidth}{!}{
\begin{tabular}{|l|l|l|l|l|l|l|}
\hline
\multirow{2}{*}{\textbf{\# of steps}} & \multicolumn{3}{l|}{\textbf{Model based}} & \multicolumn{3}{l|}{\textbf{Model free}} \\ \cline{2-7}
 & \textbf{Acc.} & \textbf{Prec.} & \textbf{Rec.} & \textbf{Acc.} & \textbf{Prec.} & \textbf{Rec.} \\ \hline
\textbf{500} & 78\% & 69.4\% & 100\% & 57\% & 53\% & 98\% \\ \hline
\textbf{750} & 85\% & 80\% & 92\% & 55\% & 52\% & 100\% \\ \hline
\textbf{1000} & 89\% & 81\% & 100\% & 78\% & 70\% & 96\% \\ \hline
\textbf{2000} & 100\% & 100\% & 100\% & 86\% & 78\% & 100\% \\ \hline
\end{tabular}
}
\caption{Blocker performance for different human intervention steps in Island Navigation. Accuracy (Acc.), Precision (Prec.), and Recall (Rec.)}
\label{fig:blocker_pref}
\end{table}

\begin{figure*}[t]
  \centering
  \subfigure[4x4 GridWorld]{\includegraphics[scale=0.5]{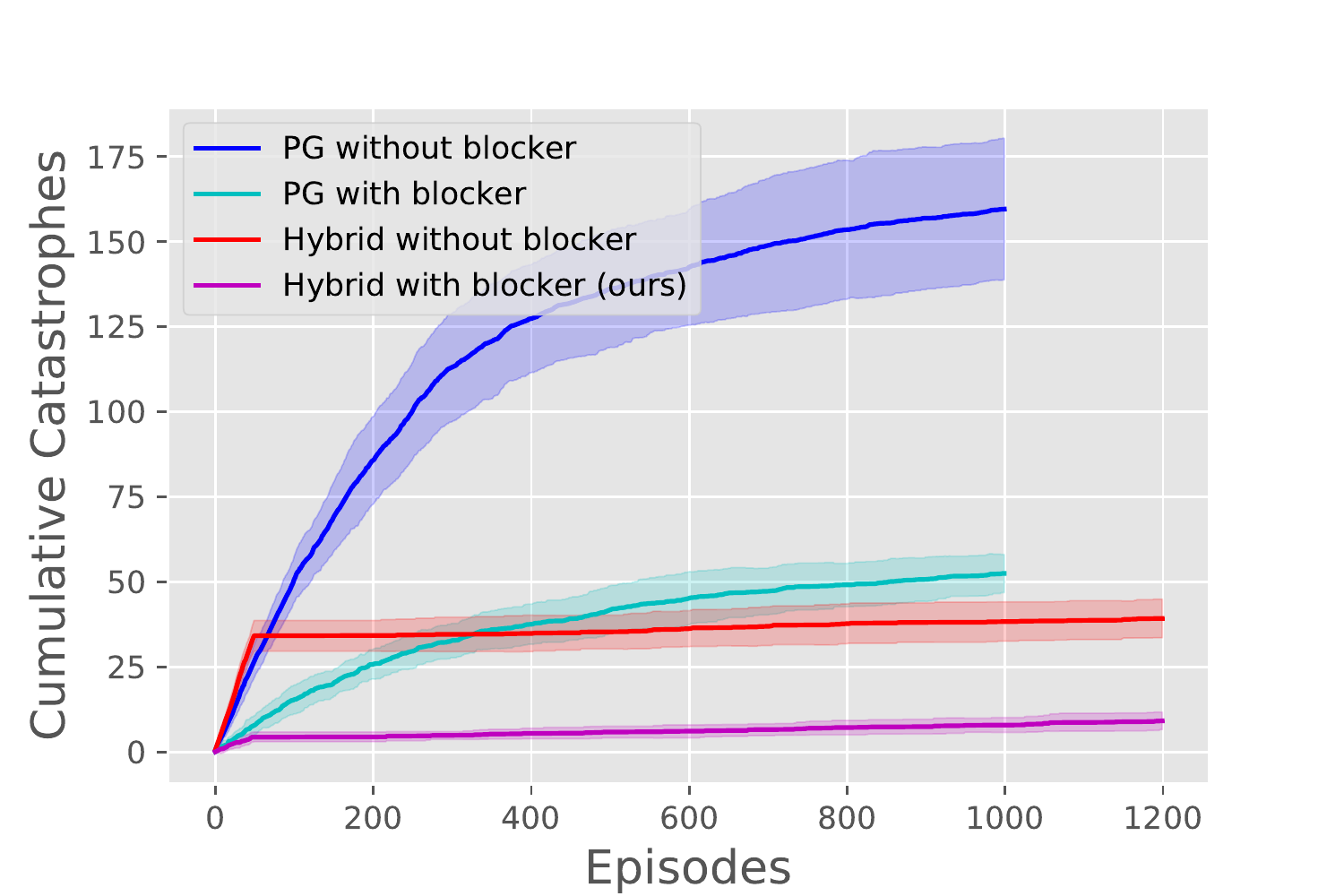}}\quad
  \subfigure[Island Navigation]{\includegraphics[scale=0.5]{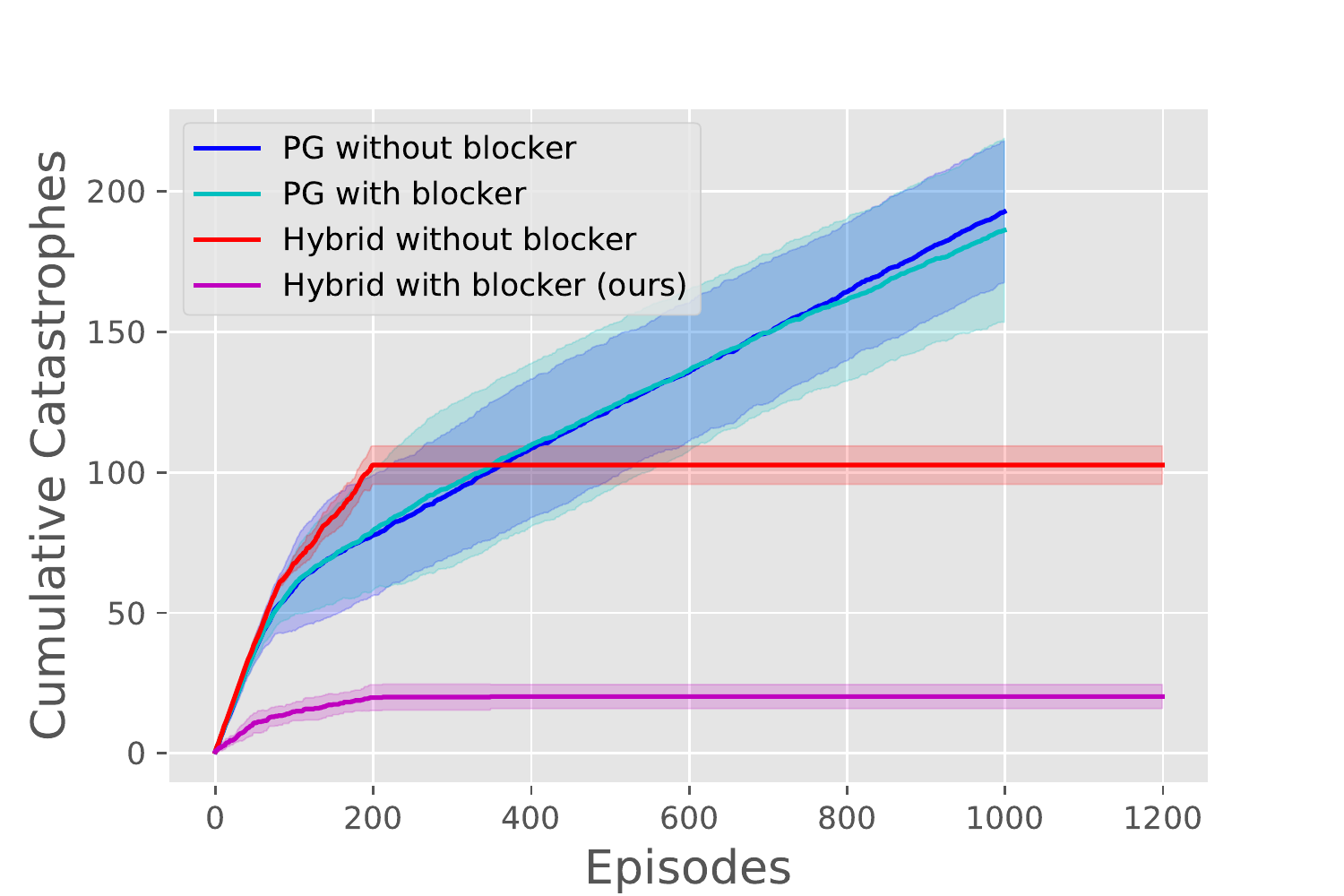}}\quad
    \caption{Cumulative Catastrophes}
    \label{fig:cum-cata}

  \centering
  \subfigure[4x4 GridWorld]{\includegraphics[scale=0.5]{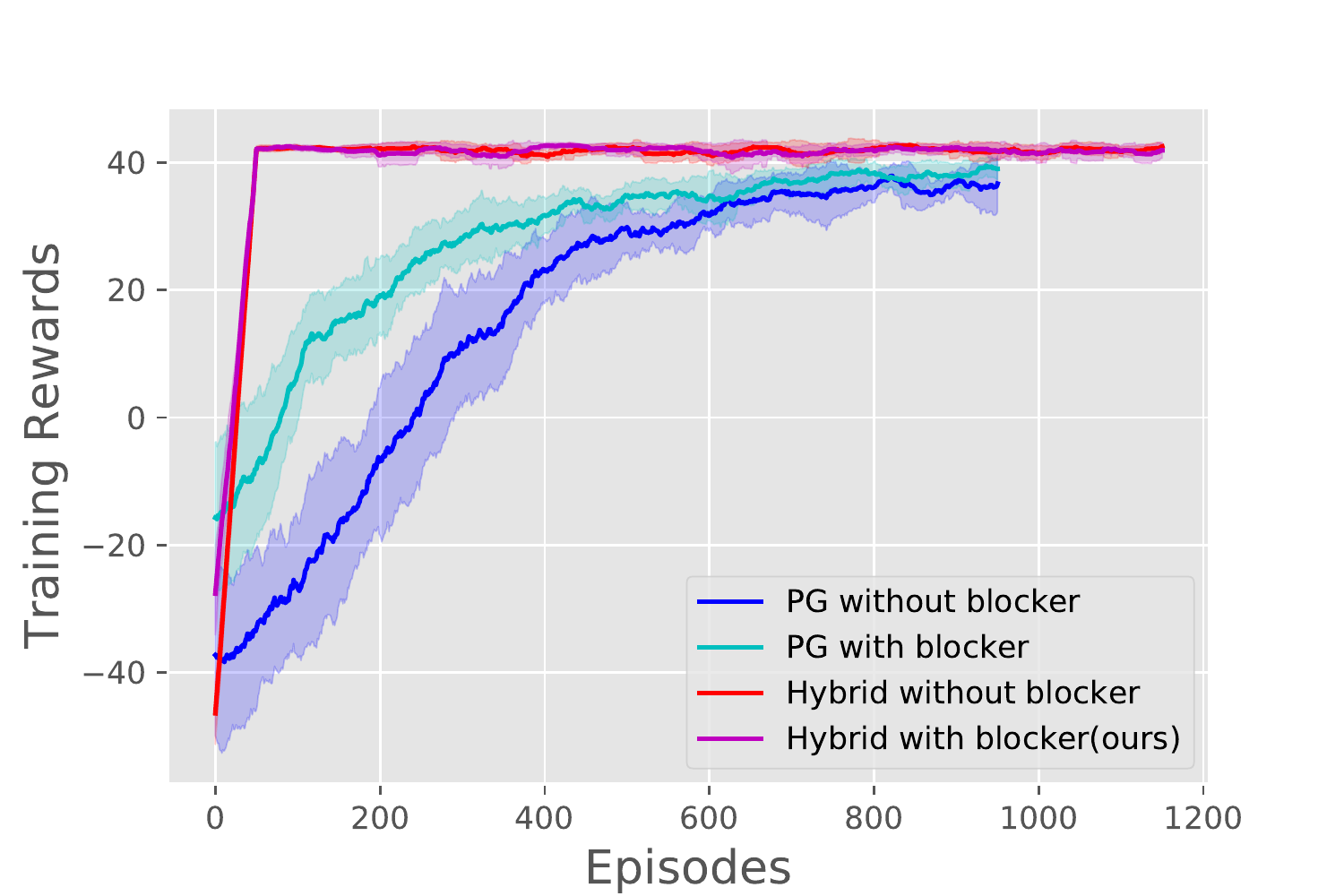}}\quad
  \subfigure[Island Navigation]{\includegraphics[scale=0.5]{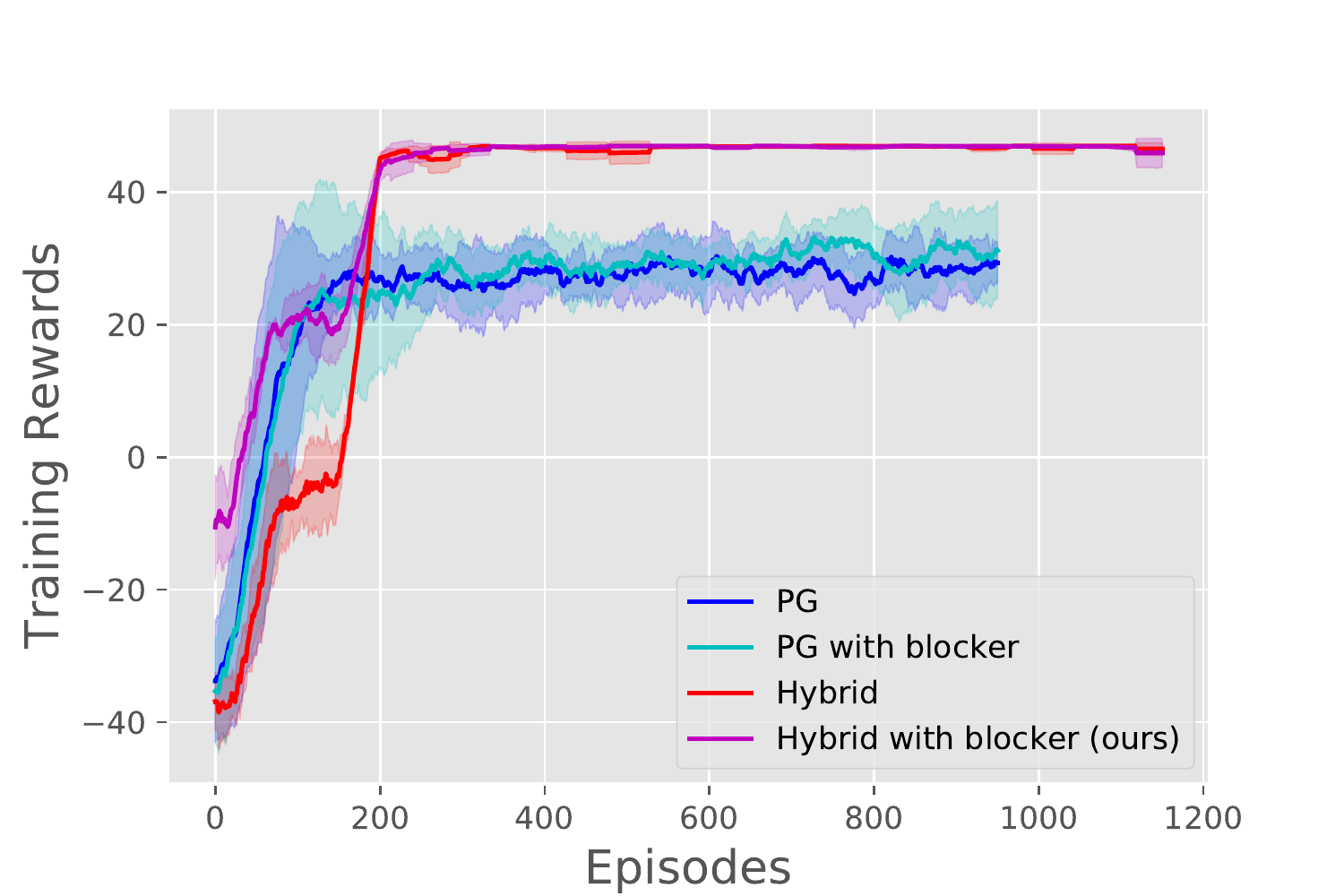}}\quad
    \caption{Evaluation Reward}
    \label{fig:rew}
\end{figure*}

\section{Experiments and Results}

We evaluate our hybrid model on two safe-RL environments with different complexities (in terms of state-space size). The first is an OpenAI Gym implementation of text based Grid-world \cite{brockman2016openai} and the second is Deepmind's AI safety Grid-world \cite{Leike}. Using these two environments, we are able to perform our experiments on both the low-dimensional standard representation (Figure \ref{fig:env} (a)) as well as the higher-dimensional visual representation (Figure \ref{fig:env} (b)). Using these two different environmental representations will allow us to test our method in not only simple tasks but also in harder, more complex tasks involving higher-dimensional state-spaces.

We evaluate and compare our models using two metrics: \emph{Cumulative Catastrophes} (i.e. the number of times agent goes to an unwanted state during the training phase) and \emph{Rewards} (the total amount of reward gathered by the agent during each episode). We compare our model to a traditional RL agent using the REINFORCE policy gradient algorithm \cite{sutton2000policy} which is one example of a model free algorithm. Additionally, for both our hybrid approach, as well as the policy gradient approach, we evaluate their effectiveness with and without learning in the presence of a trained blocker agent. This allows us to get a sense of the sample efficiency gains from combining model-based learning with model-free learning as well as the performance gains from training a blocker agent using model-based policies.

\subsection{4x4 GridWorld}
In this case study we used a text based toy environment from OpenAI gym environments \cite{brockman2016openai}. This environment consists of a standard representation with $ S \in R^{16}$ states, $A \in \{up, down, left, right\}$ actions corresponding to moving up, down, left and right. As shown in Figure \ref{fig:env} (a) a blue square denotes the position of agent, the goal state is denoted by the green and the red square signifies a fire (catastrophic) state. The goal of the agent is simply to navigate from the start state to the goal state as quickly as possible while avoiding the fire states.

\subsection{Island Navigation}
Figure \ref{fig:env} (b) shows our second environment called Island Navigation, which is from Deepmind's AI safety Grid-world environment. This environment uses a visual representation with $ S \in R^{1024}$ states, $A \in \{up, down, left, right\}$ actions. In this environment, the light blue square represents the agent, while the blue blocks, which represent water (catastrophic), should be avoided by agent. The Goal of this agent is similar to previous environment which is to reach the green square as quickly as possible. This environment gives its observation in visual representation in the form or images with a resolution of 32 X 32 pixels.

\subsection{Blocker Performance}
We tested the performance of two different blocker agents trained on the island navigation: one with data collected from a model-free policy (similar to \citeauthor{Saunders2017}) and one with data collected from a model-based policy (our approach). Additionally, we tested both blocker agents with different amount of training data (500, 750, 1000, and 2000) from the human.
In both approaches, the blocker agents are trained using a certain number of steps during which the human is used to oversee the actions of the agent (from either a model-based policy or a model-free policy). In this way, we can get a sense of the sample efficiency of each approach by looking at blocker prediction performance (i.e. the accuracy of the blocker making the same intervention as the human) using different amounts of training data.

Table \ref{fig:blocker_pref} shows the performance of the blocker with the various human intervention steps. For each training set size, the blocker was evaluated on a held out test set to measure the performance in terms of accuracy, precision and recall. As can be seen from Table~\ref{fig:blocker_pref}, the model-based blocker agent achieves on average 20\% higher accuracy performance than the model-free blocker. More importantly, is the recall, which represents the ratio of true-positives to true-positive + false negatives. A recall of 100\% is very important since it indicates 0 false negatives which would mean that a blocker never missed the blocking of a bad action. Here, we see that the model-based approach achieves 100\% in 3 of the four tested training sizes and on average achieves a higher recall than the model-free blocker. This increased performance is likely due to the model-based blocker seeing a much better distribution of data when we train it during the model based system since the model-free approach explores less randomly than the model-based approach.
Hence the model based agent is able to train a much more robust blocker which is important for safe exploration.

\subsection{Safe-RL performance}
In this section we show the performance of our hybrid model-based and model-free approach to safe RL using a blocker agent trained from human intervention examples. We demonstrate the performance of our approach in terms of the cumulative catastrophes (i.e. the total number of time the agent went into a catastrophic or failure state), as well as the total amount of environment reward received. We compare our method against a traditional model-free approach using a policy gradient algorithm. Additionally, we test the improvement in safety and performance gained from the blocker agent by tested our method and the model-free approach both with and without a blocker agent. It's important to note that for these experiments, we used blocker agents trained only to 1000 steps, in which the blocker agents are not yet at 100\% accuracy. We do this so that we get a better comparison of the gains in terms of safety that would be harder to access than if we used a perfect blocker agent.

\subsubsection{Cumulative Catastrophes}
Figures~\ref{fig:cum-cata}a and \ref{fig:cum-cata}b show the cumulative catastrophes encountered during training of four comparison methods: the policy gradient method (PG), the policy gradient method trained with a blocker (PG with blocker), our hybrid approach without a blocker (Hybrid) and our hybrid approach trained with a blocker (Hybrid with blocker (ours)).  In both tasks (4x4 GridWorld and Island Navigation) we see that even without a blocker, our model-based, hybrid approach encounters less catastrophic states than the model-free, policy gradient approach (49 catastrophic states compared to 162 for 4x4 GridWorld and 100 compared to 188 for Island Navigation). Additionally, we see that when trained with a blocker agent, our hybrid approach with a blocker encounters only 7 catastrophes in the 4x4 GridWorld environment and only 22 catastrophes in Island navigation environment, which is significantly less than the policy gradient with a blocker which encounters 54 catastrophic states in 4x4 Gridworld and 157 in Island Navigation.

\subsubsection{Rewards}
\par Figures \ref{fig:rew}a and \ref{fig:rew}b show the performance comparison of the four methods in terms of total reward obtained during training, which give a sense of the quality of the learned policy as well as the number of samples required to reach that policy.
For the PG and PG with blocker conditions, the model-free agent was trained for 1200 episodes in total. For our hybrid approach (both with and without a blocker agent), the dynamics model is trained and used with the MPC controller during the first 200 episodes. Afterwards, our system switches to model-free learning (bootstrapped using data from MPC (Figure \ref{arch})) and trains for 1000 episodes (1200 episodes total).

As can be seen in Figure \ref{fig:rew}, The hybrid approach is able to achieve the maximum task reward (i.e. perfect task completion) for both the 4x4 GridWorld and Island Navigation environments when both learning with and without the blocker (achieving a reward of 45 and 47 for both environments respectively). The policy gradient approach, however, was not able to achieve the maximum reward for either environment. The influence of the blocker agent can be seen in terms of the speed of convergence for each model (i.e. models trained with the blocker agents trained faster than when training without the blocker). This coupled with the results from Figure \ref{fig:cum-cata} show that the blocker allows policies to not only train faster but train safer.

\section{Discussion and Conclusion}

We presented a hybrid architecture to improve sample efficiency and reduce the amount of human time required to ensure safe training of RL agents. We show that the blocker trained during the model based system works better than the model-free approach. We also show that our hybrid architecture, a combination of model-based, MPC and model-free methods, is more data efficient than standard model-free approaches, allowing the agent to reach a stable policy in a fast and efficient manner.

Similar to some previous work, our blocker agent is trained to imitate the task of human intervention and ensure safe exploration. This is necessary because as humans cannot be present during the entire training phase of the RL agent (which is often very long), the idea is to handover this task to a trained blocker that will intervene on the human's behalf and thus will continue to allow for safe training of the RL agent.
The blocker agent, however, may not be always perfect since it is very difficult to train it to block every unsafe action from all possible states (especially in states that the RL agent has never encountered before). This problem is also discussed in \cite{Saunders2017} where they follow a similar process. In this paper, we tried to mitigate this problem in two ways. First, the dataset for the blocker was collected during the model based training phase. We showed that this is better than collecting the dataset during a typical model-free training cycle. This ensured that the quality of the blocker agent was better. Second, and more important, we used a combination of model-based and model free system which results in faster training and increased sample efficiency \cite{nagabandi2018neural}. In this way, the agents reach a stable state with less data and will likely learn to avoid bad states much faster. This in turn means that the agent will act safely, even when the blocker fails to intervene.

Another benefit from our method is that we can potentially use the trained model based system to quickly train RL agents to perform completely new tasks since our learned dynamics model and blocker agent are task independent.

We can formulate a new reward function in the MPC and initialize a model-free system to learn this new task. For example, suppose a robotic agent is being trained to navigate a room and perform a task and the blocker is trained so that the robot never knocks over and breaks any objects (unsafe) even while learning and exploring. The same blocker and environment model can now be used to train a completely different task which still requires the robot to interact with objects in a safe manner as long as the environment and agent dynamics remain the same. This will be explored further in future work.

Even though we performed experiments on standard as well as visual representations, the environments tested in this paper are still relatively small environments with simple state and action spaces. The next step would be to explore how these methods perform on more complex higher dimensional state and action spaces. Overall, we believe our method demonstrates the importance of incorporating model-based and model-free approaches with human-interaction for training safe RL agents.

\section{ Acknowledgments}
This project was sponsored by the U.S. Army Research Laboratory under Cooperative Agreement Number W911NF-10-2-0022. The views and conclusions contained in this document are those of the authors and should not be interpreted as representing the official policies, either expressed or implied, of the U.S. Government. The U.S. Government is authorized to reproduce and distribute reprints for Government purposes notwithstanding any copyright notation herein.

\bibliography{ref.bbl}
\bibliographystyle{aaai}
\end{document}